\documentclass{article}
\usepackage{spconf,amsmath,graphicx}

\usepackage{xcolor}
\usepackage{subcaption}
\usepackage{multirow}
\usepackage{amssymb}

\title{Longer Version for ``Deep Context-Encoding Network for Retinal Image Captioning''}
\name{Jia-Hong Huang$^{1}$, Ting-Wei Wu$^{2}$, Chao-Han Huck Yang$^{2}$, Marcel Worring$^{1}$ }
\address{ $^{1}$University of Amsterdam,  Netherlands \\ $^{2}$Georgia Institute of Technology, USA }
\begin{document}
%
\maketitle
\begin{abstract}
Automatically generating medical reports for retinal images is one of the promising ways to help ophthalmologists reduce their workload and improve work efficiency. In this work, we propose a new context-driven encoding network to automatically generate medical reports for retinal images. The proposed model is mainly composed of a multi-modal input encoder and a fused-feature decoder.
Our experimental results show that our proposed method is capable of effectively leveraging the interactive information between the input image and context, i.e., keywords in our case. The proposed method creates more accurate and meaningful reports for retinal images than baseline models and achieves state-of-the-art performance. This performance is shown in several commonly used metrics for the medical report generation task: BLEU-avg (+16\%), CIDEr (+10.2\%), and ROUGE (+8.6\%).
\end{abstract}
\section{Introduction}

According to \cite{huang2021deepOpht,pizzarello2004vision}, retinal diseases, e.g., Diabetic Retinopathy (DR) and Age-related Macular Degeneration (AMD) are expected to affect over 500 million people worldwide. So, the workload of ophthalmologists will be overwhelming. Automating part of the retinal disease diagnosis procedure \cite{huang2021deepOpht}, such as medical report generation for retinal images, is one of the good ways to help them reduce the workload.

The authors of \cite{laserson2018textray,jing2018automatic,li2018hybrid,yang2021pate,yang2020characterizing} have proposed methods to cope with the automatic medical report generation. These proposed approaches work on image content only because they are mainly based on traditional natural image captioning models \cite{xu2015show,karpathy2015deep}. However, it is hard to generate abstract medical concepts or descriptions, \cite{laserson2018textray,jing2018automatic}, i.e., key components of medical reports \cite{huang2021deepOpht}, only based on image information. To address this issue, the authors of \cite{huang2021deepOpht,huang2021deep} have proposed a context-driven, i.e., in the form of keywords sequence, medical report generation method for retinal images. Since the context-driven method has multi-modal inputs, i.e., the keywords and image, the authors of \cite{huang2021deepOpht} exploit the average method to fuse the multi-modal information. However, fusing the multi-modal information by the average method in this case probably cannot effectively capture the interactive information between the context and image \cite{yang2020characterizing,yang2021pate,huang2019novel,huang2020query,huang2021contextualized,huang2021gpt2mvs,huang2017robustness,huang2017vqabq,huang2018robustness,liu2018synthesizing,yang2018auto,yang2018novel,hu2019silco,huang2019assessing,agrawal2017vqa,huang2021deepOpht}. 
According to \cite{jing2018automatic,agrawal2017vqa}, such interactive information can largely affect the quality of the generated descriptions.

In this work, we propose an effective approach for capturing the interactive information for context-driven medical report generation for retinal images. We assume that each word of the input context, i.e., keywords sequence \cite{huang2021deepOpht} in our case, has different levels of interactive effects to the image information. We claim that if we can better capture such effects/information, the model performance can be improved. So, to better capture the interactive information, in our proposed method, we exploit the LSTM-based structure \cite{hochreiter1997long} to encode the interactive information between the image and keywords. Based on the experimental results, our proposed method is capable of effectively capturing the interactive information between the keywords and image. Also, it generates more accurate and meaningful descriptions for retinal images. In this paper, our contributions are summarized as follows: 

\vspace{+4pt}
\noindent\textbf{Contributions.}
\begin{itemize}
    \item We propose a new end-to-end encoder-decoder model for context-driven medical report generation for retinal images and the model achieves the state-of-the-art performance under popular evaluation metrics for medical report generation, i.e., BLEU, CIDEr, and ROUGE. 
    
    \item We conduct several experiments to show that the proposed method is capable of effectively leveraging the interactive information between the input image and input keywords.

\end{itemize}

\section{Related Work}

In this section, since most medical report generation models are based on natural image captioning models, we first review the existing works of image captioning. Then, we discuss the works related to context-driven medical report generation.

\noindent\vspace{+0.1cm}\textbf{2.1 Image Captioning}

The authors of \cite{vinyals2015show,karpathy2015deep,fang2015captions} have proposed a new computer vision task, i.e., image captioning. The goal of this task is to generate text-based descriptions for a given image. The authors of \cite{vinyals2015show} have proposed an encoder-decoder based image captioning model. They exploit a convolutional neural network (CNN) model, which is considered as an encoder, to extract the image feature. Then, they use the extracted image feature as an input at the first time step of an RNN model, considered as a decoder, to generate the description of a given input image. The authors of \cite{karpathy2015deep} have proposed a model that is able to embed visual and language information into a common space. In \cite{fang2015captions}, the authors use a language model to combine a set of possible words, which are related to several parts of a given input image, and then generate the description of the image. In \cite{gao2019deliberate}, the authors have proposed a deliberate residual attention image captioning network. They exploit the layer of first-pass residual-based attention to generate the hidden states and visual attention for creating a preliminary version of the image descriptions, while the layer of second-pass deliberate residual-based attention refines them. The authors claim that their proposed method has the potentials to generate better captions because the second-pass is based on the global features captured by the visual attention and hidden layer in the first-pass. The authors of \cite{pan2020x} introduce a unified attention block that fully employs bilinear pooling to selectively capitalize on visual information or perform reasoning. In \cite{hendricks2016generating}, the authors have introduced an approach focusing on discriminating properties of the visible object, jointly predicting a class label and explaining why the predicted label is proper for a given input image. Through a loss function based on reinforcement learning and sampling, their proposed model learns to generate captions for the given image. The authors of \cite{liu2017improved} mention most existing image captioning models are trained via maximum likelihood estimation. However, the log-likelihood score of some description cannot correlate well with human assessments of quality. Standard syntactic text evaluation metrics, e.g.,  BLEU \cite{papineni2002bleu} and ROUGE \cite{lin2004rouge}, are also not well correlated. In \cite{liu2017improved}, the authors have shown how to exploit a policy gradient method to optimize a linear combination of CIDEr \cite{vedantam2015cider} and SPICE \cite{anderson2016spice}. We observe that most existing image captioning models work well on natural image dataset, but often do not generalize well to medical image datasets. So, we need a specialized method, e.g., through context-driven, for medical report generation.

\noindent\vspace{+0.1cm}\textbf{2.2 Context-driven Medical Report Generation}

The authors of \cite{malinowski2015ask,agrawal2017vqa} have proposed models for the visual question answering (VQA) task. The goal of this task is to answer, i.e., outputting an answer, a given input question based on a given input image. Since the VQA task has visual and textual inputs, VQA models exploit one modality to help the other, i.e., context-driven, \cite{agrawal2017vqa}. A similar idea of multiple input modalities can be applied to medical report generation \cite{jing2018automatic,huang2021deepOpht,huang2021deep}. In \cite{jing2018automatic}, the authors exploit an image input to generate intermediate/side products, i.e., text-based tags, to help the later generation of the medical report. Note that the proposed model \cite{jing2018automatic} only has an image input, i.e., single modality. The authors of \cite{huang2021deepOpht} have proposed a model that can exploit multi-modal inputs, i.e., the expert-defined keywords and image, to improve the model performance and generate a better quality of medical reports. The main difference between text-based tags \cite{jing2018automatic} and keywords \cite{huang2021deepOpht} is that text-based tags are model-generated intermediate products which could be wrong/bias information. The intermediate products could confuse models during the training phase. However, keywords are expert-defined information and the correctness of the information is guaranteed by experts. So, it helps models learn better \cite{huang2021deepOpht}. Although multi-modal information improves model performance, the fusion of multi-modal inputs will become another challenging issue. In this work, we propose a model with a better fusion mechanism to address this issue.

\section{Methodology}
In this section, we present our medical report generation model and illustrate methods about how we train the model with knowledge of keywords. We introduce a multi-modal input encoder to extract information both from images and keywords inputs and use the encoder-decoder architecture for the medical report generation task.

\noindent\vspace{+0.1cm}\textbf{3.1. Multi-modal Input Encoder $E_k$}

\begin{figure}[t!]
\centering
\includegraphics[width=0.8\linewidth]{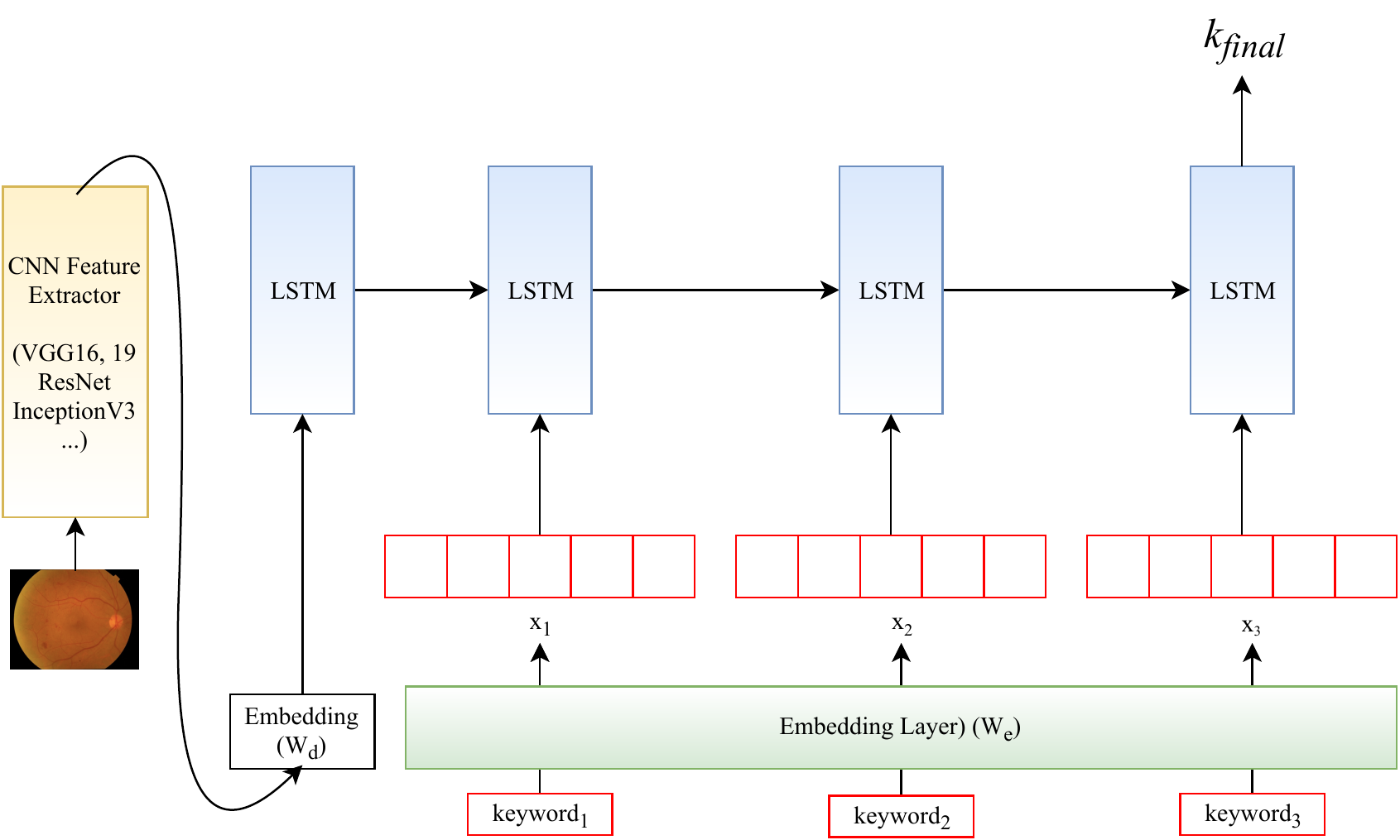}
\vspace{-0.3cm}
\caption{The figure sketches how our \textit{Multi-modal Input Encoder} works for image and keyword embedding. The image content is introduced in the first time step of an LSTM, with keyword vectors following behind. The encoder will generate an image-keyword fused vector to feed back in the structure shown in Figure \ref{fig:figure2}.}
\label{fig:figure1}
\end{figure}

In \cite{huang2021deepOpht}, the authors exploit an intuitive mechanism, i.e., the average method, to fuse features. Although the average mechanism seems straightforward, somehow loses different levels of interactive information with the image contents. Therefore, in this paper, we introduce an LSTM-based structure, , referring to Figure \ref{fig:figure1}, to better capture the interactive information. The image $I$ is input once at $t=-1$ to inform LSTM about the image contents. See Equation-(\ref{eq:1}).

\begin{equation}
    x_{t} = W_d\times \phi(I), t=-1
	\label{eq:1}
\end{equation}

Then, we feed each keyword embedded vector ${x_n}$ to keep LSTM in memory. See Equation-(\ref{eq:2}).

\begin{equation}
    x_n = W_kk_n, n \in \{0,...,N\}
	\label{eq:2}
\end{equation}

\begin{figure}[t!]
\includegraphics[width=1.0\linewidth]{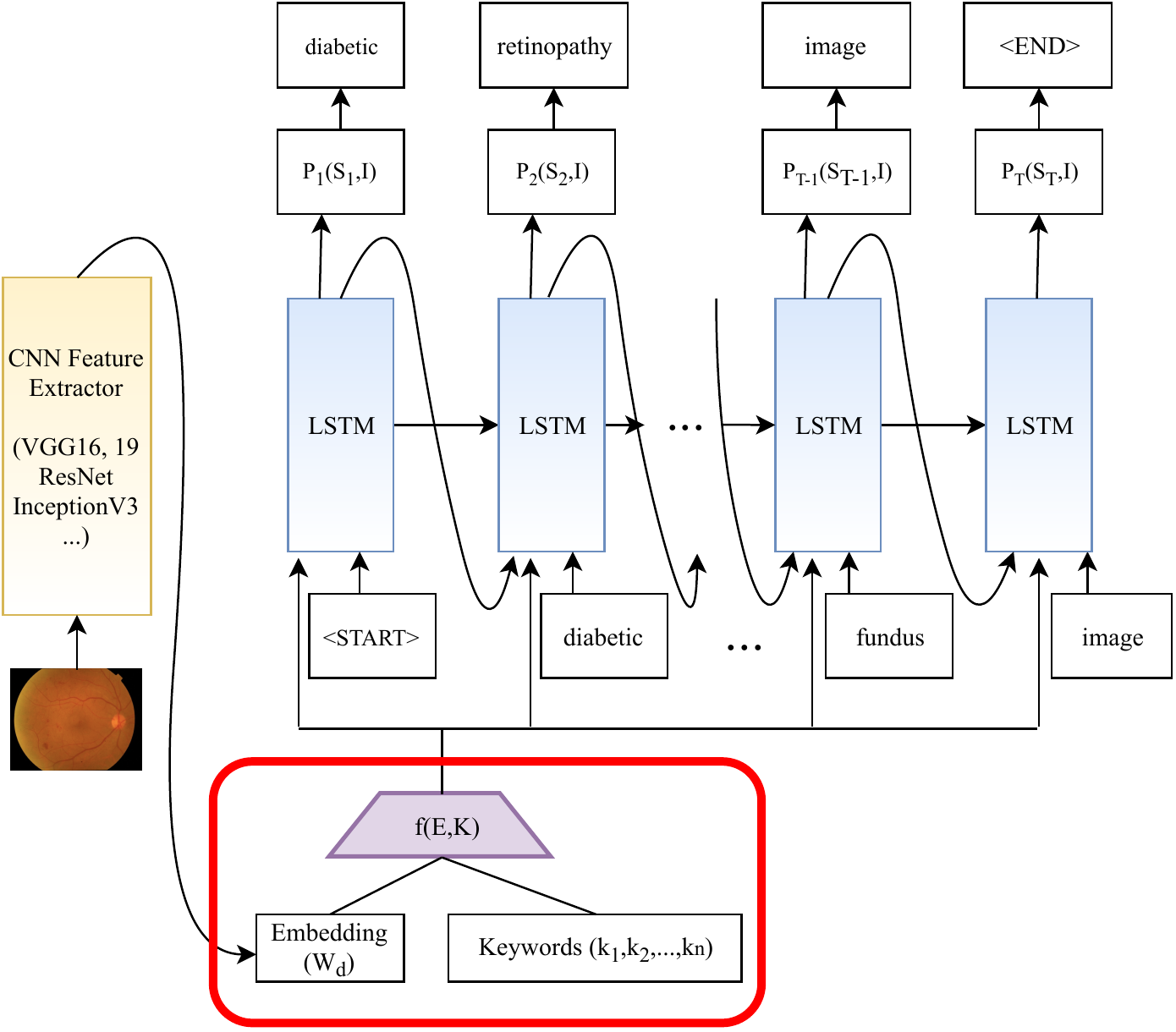}
\vspace{-0.5cm}
\caption{The figure shows our \textit{Fused-feature Decoder}. Before fed in the LSTM, the image content is preprocessed with keyword embedded vectors by \textit{Multi-modal Input Encoder} $E_k$.}
\label{fig:figure2}
\end{figure}

Finally, we extract the last hidden state $k_{final}$, referring to Equation-(\ref{eq:3}), for our final fused vector to feed it back into our fused-feature decoder, referring to Figure \ref{fig:figure2} and the next subsection.

\begin{equation}
    k_{final} = LSTM(x_t), t \in \{0,...,N-1\}
	\label{eq:3}
\end{equation}

\noindent\vspace{+0.1cm}\textbf{3.2. Fused-feature Decoder $G_\theta$}

For our fused-feature decoder, i.e., the description generator, we adopt the same CNN image embedder $\phi$, used in \cite{huang2021deepOpht}, to extract image features and feed in each time step of a subsequent bidirectional LSTM model, and all preceding words as defined by $p(S_t|I, S_0,..., S_t-1)$. We denote a true sentence describing the input image as $S=(S_0,..., S_T)$. Then, we unroll our description generator as shown in Equation-(\ref{eq:4}), (\ref{eq:5}), (\ref{eq:6}), and (\ref{eq:7}). In Equation-(\ref{eq:4}) and (\ref{eq:5}), we represent each word as a bag-of-word id $S_t$. The words $S$ and image vector $I$ are mapped to the same space: the image by using an image encoder $\phi$, i.e., a deep CNN, a fully-connected layer $W_d \in \mathbb{R}^{E \times F}$ and the words by word embedding $W_e \in \mathbb{R}^{E \times V}$. $E$ represents the word embedding size, $F$ is the image feature size, and $V$ is the number of all vocabulary in captions. 

\begin{equation}
    e_t = W_d\times \phi(I), t \in \{0,...,T\}
	\label{eq:4}
\end{equation}
\begin{equation}
    x_t = W_eS_t, t \in \{0,...,T\}
	\label{eq:5}
\end{equation}

In Equation-(\ref{eq:6}) and (\ref{eq:7}), for each time step, we feed the network with image contents $e_t$, fused multi-modal feature $k_{final}$, and ground truth word vector $x_t$ to strengthen its memory of images. We also exploit the dropout technique to alleviate the effect of overfitting and noises. Finally, we denote $P_I$ as the true medical descriptions for $I$ provided in the training set and $P(S, I)$ as the final probability distribution after one fully-connected layer and softmax function. The loss function $L(P|I,S)$ is calculated as the sum of the negative log-likelihood at each time step.

\begin{equation}
    P_{t} = BiLSTM([e_t, k_{final}, x_t]), t \in \{0,...,T\}
    \label{eq:6}
\end{equation}
\begin{equation}
    L(P|I,S) = \mathbb{E}_{S\sim P_I}{[log{P(S,I)]}}
    \label{eq:7}
\end{equation}

For the inference phase, we exploit \textit{Beam Search} to generate a sentence given an image. We consider the set of $k$ sentences up to time step $t$ to be candidates and generate $P_{t+1}$, and take the best $k$ sentences. Note that $k$, i.e., a user-specified parameter, indicates number of beams, e.g., $k=3$.

\section{Experiments and Analysis}
In this section, we evaluate and analyze the effectiveness of our proposed method based on the same assumption, made by \cite{huang2021deepOpht,jing2018automatic}, that an effective deep model is helpful in practice.   

\vspace{+1pt}\noindent\textbf{4.1. Dataset, Experimental Setup, and Evaluation Metrics}

The authors of \cite{huang2021deepOpht} have proposed a state-of-the-art model for medical report generation for retinal images. Also, they introduce a large-scale retinal image dataset with unique expert-defined keyword annotations. The keywords annotations contain important information about patients and potential diseases based on retinal image analysis and conversation with the patients. The dataset is composed of 1,811 grey-scale Fluorescein Angiography (FA) images and 13,898 colorful Color Fundus Photography (CFP) images. In the dataset, each retinal image has two corresponding labels, i.e., clinical description and keywords. The word length in the dataset is between 5 and 10 words. We follow the original setup of the dataset, i.e., $60\%/20\%/20\%$ for training/validation/testing, respectively. Note that we take the keywords label and retinal image as our inputs and clinical description as our ground truth prediction.

Similar to \cite{huang2021deepOpht}, we adopt image feature extractors $\phi$, pre-trained on ImageNet, to extract our retinal image features. The layer before the last fully-connected layer is used for embedding features that are ready to feed into the decoder. To process the annotations and keywords in the DEN dataset, non-alphabet characters are removed, all remaining characters are converted to lower-case, and all the words appearing only once are replaced by a special token $<UNK>$. Then, our vocabulary size $V=4007$ when keywords are excluded and $V_{k}=4292$ when keywords are included. Our sentences are truncated or padded with a max length of $50$. For the word embedding layer, we use an embedding size of $E=300$ to encode words, and a hidden layer size $H_{LSTM}=256$. We set the learning rate to $0.001$ to train the models with two epochs and the mini-batch size to $64$. 

Finally, in our experiments, we exploit the same evaluation metrics, used in \cite{huang2021deepOpht}, for medical report generation, i.e., \cite{papineni2002bleu,lin2004rouge,vedantam2015cider}, to evaluate our model performance.

\begin{table*}[t!]
\caption{This table shows the evaluation results of the keyword-driven and non-keyword-driven models. Note that we highlight the best scores of keyword-driven and non-keyword-driven generators in each column, respectively. ``w/o'' denotes non-keyword-driven models, and ``w/'' denotes keyword-driven models. ``BLEU-avg'' denotes the average score of BLEU-1, BLEU2, BLEU-3, and BLEU-4. Based on this table, we find that all the keyword-driven models, with our proposed feature fusion method, are superior to the non-keyword-driven models.}
\centering
\scalebox{1.0}{
\begin{tabular}{|c|l|l|l|l|l|l|l|l|}
\hline
\multicolumn{2}{|c|}{Model}                  & BLEU-1 & BLEU-2 & BLEU-3 & BLEU-4 & BLEU-avg & CIDEr & ROUGE \\ \hline
\multirow{2}{*}{Karpathy, et al. \cite{karpathy2015deep}}           & w/o & 0.081  & 0.031  & 0.009  & 0.004  & 0.031    & 0.117 & 0.134 \\ \cline{2-9}
                                    
                                    & w/ &  \textbf{0.219}     & \textbf{0.134}   & \textbf{0.074}      & \textbf{0.035}      & \textbf{0.116}      & \textbf{0.398}            & \textbf{0.252} \\ \hline

\multirow{2}{*}{Vinyals, et al. \cite{vinyals2015show}}              & w/o & 0.054  & 0.018  & 0.002  & 0.001  & 0.019    & 0.056 & 0.083 \\ \cline{2-9} 
                                    
                                    & w/ &  \textbf{0.156}     & \textbf{0.088}   & \textbf{0.042}      & \textbf{0.016}      & \textbf{0.076}      & \textbf{0.312}            & \textbf{0.200} \\ \hline

\multirow{2}{*}{Jing, et al. \cite{jing2018automatic}}              & w/o & 0.130  & 0.083  & 0.044  & 0.012  & 0.067    & 0.167 & 0.149 \\ \cline{2-9} 
      
                                    & w/ &  \textbf{0.216}     & \textbf{0.131}      & \textbf{0.075}       & \textbf{0.037}    & \textbf{0.115}     & \textbf{0.385}            & \textbf{0.258} \\ \hline

\multirow{2}{*}{Li, et al. \cite{li2019knowledge}}        & w/o & 0.111  & 0.060  & 0.026  & 0.006  & 0.051    & 0.066 & 0.129 \\ \cline{2-9} 

                                    & w/ & \textbf{0.217}    & \textbf{0.139}  & \textbf{0.079}   & \textbf{0.043}   & \textbf{0.120}   & \textbf{0.525}   & \textbf{0.267} \\ \hline

\end{tabular}}
\label{table:table200}
\end{table*}

\begin{table*}[t!]
\caption{This table is to show that our proposed method performs better than baseline models under the ``Image + Keywords'' situation. Note that ``mul'' denotes element-wise multiplication, and ``sum'' denotes summation. The results are based on the keyword-driven model \cite{karpathy2015deep} in Table \ref{table:table200}.}
\centering
\scalebox{1.0}{
\begin{tabular}{|c|l|l|l|l|l|l|l|l|}
\hline
\multicolumn{2}{|c|}{Fusing method}     & BLEU-1 & BLEU-2 & BLEU-3 & BLEU-4 & BLEU-avg & CIDEr & ROUGE \\ \hline
\multicolumn{2}{|c|}{Baseline-1 (sum)}  & 0.014  & 0.002  & 0.001  & 0.000  & 0.004    & 0.019 & 0.023 \\ \hline
\multicolumn{2}{|c|}{Baseline-2 (mul)}  & 0.077  & 0.031  & 0.004  & 0.001  & 0.028    & 0.042 & 0.102  \\ \hline
\multicolumn{2}{|c|}{DeepOpht \cite{huang2021deepOpht}}  & 0.184  & 0.114  & 0.068  & 0.032  & 0.100    & 0.361 & 0.232  \\ \hline
\multicolumn{2}{|c|}{Our method} & \textbf{0.219}  & \textbf{0.134}   & \textbf{0.074}  & \textbf{0.035}  & \textbf{0.116}  & \textbf{0.398}  & \textbf{0.252} \\ \hline
\end{tabular}}
\label{table:table201}
\end{table*}

\vspace{+1pt}\noindent\textbf{4.2. Experimental Results and Effectiveness Analysis}

\vspace{+1pt}\noindent\textbf{Effectiveness of Keywords.}
In \cite{huang2021deepOpht}, the authors have shown that their proposed average-based method can exploit the keywords to help models and achieve state-of-the-art performance. We claim that our proposed method can effectively use the keywords information and achieve better performance than \cite{huang2021deepOpht}. For a fair comparison, we follow the same experimental setup, such as CNN feature extractors, etc, as mentioned in \cite{huang2021deepOpht} to conduct our experiments with the keyword-driven and non-keyword-driven models. According to Table \ref{table:table200}, the results show that all the keyword-driven models are superior to the non-keyword-driven models based on our proposed method. In Table \ref{table:table201}, the results show that our proposed method performances better than \cite{huang2021deepOpht}. The performance increases about $16$\% in BLEU-avg, $10.2$\% in CIDEr, and $8.6$\% in ROUGE. The above implies our proposed method is capable of better capturing the interactive information between the keywords and image. So, our claim is well proved.

\begin{figure*}[t!]
\includegraphics[width=0.79\linewidth]{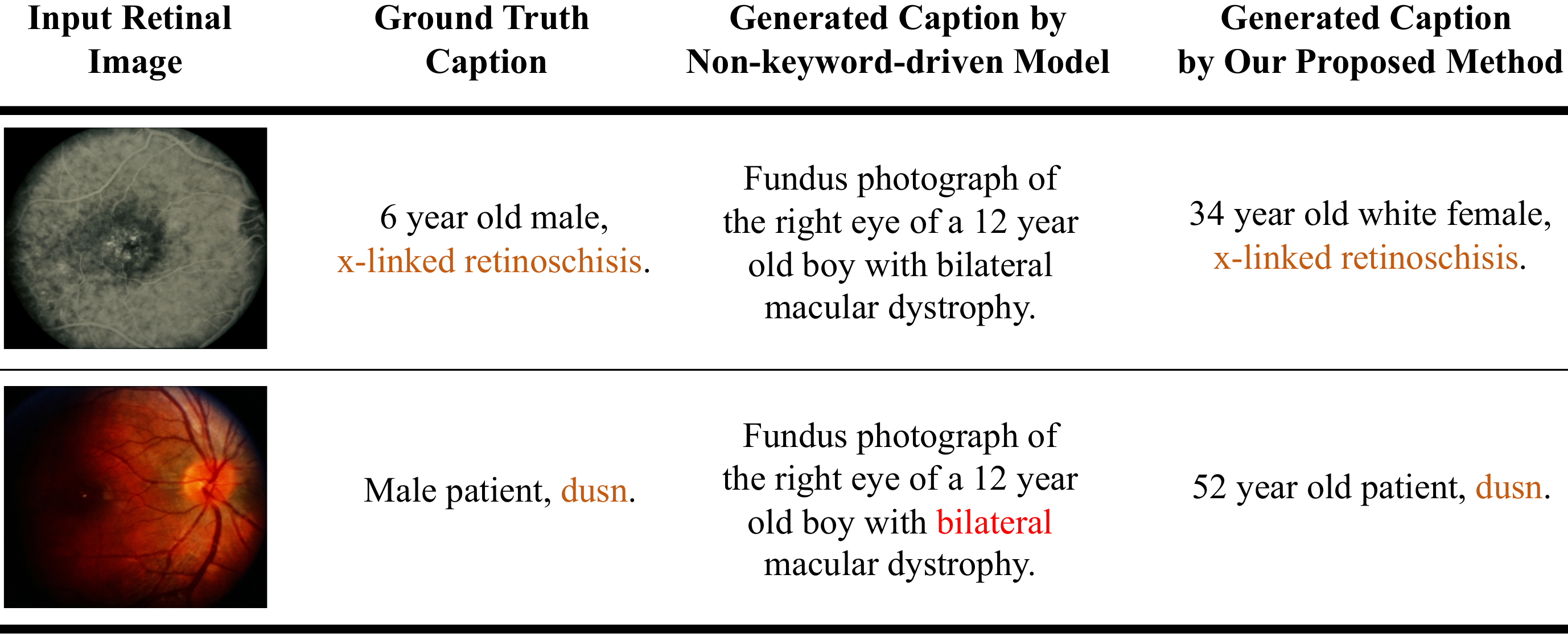}
\centering
\vspace{-0.2cm}
\caption{This figure shows some randomly selected qualitative results. Note that for the first-row example, the ground truth keywords are ``x-linked'', ``retinoschisis'', and ``xlrs''. For the second row example, the ground truth keywords are ``diffuse'', ``\textbf{unilateral}'', ``subacute'', ``neuroretinitis'', and ``dusn''. The result shows that our model effectively exploits keywords information as guidance to generate better captions. Also, the result shows that if keywords do not exist or cannot be effectively used, it is possible that: (i) For two different retinal images, models may generate the same description because of the very similar low-level features. (ii) Models may generate words with the wrong meaning, e.g., ``unilateral'' vs. ``bilateral'', referring to the second-row example and the corresponding ground truth keywords.}
\label{fig:figure3}
\end{figure*}

\vspace{+1pt}\noindent\textbf{Qualitative Results and Analysis.}
In Figure \ref{fig:figure3}, we show some qualitative results generated by our medical report generation model for retinal images. Similar to \cite{huang2021deepOpht}, our model is not capable of creating correct ``age'', ``gender'', or ``skin color'' as these are not present in the content. As mentioned \cite{huang2021deepOpht}, ideally, such ``age'', ``gender'', or ``skin color'' information would be part of the dataset, and that a system should make it part of the description by slot filling or post-processing. However, comparing to baseline models, our model can generate correct descriptions to important characteristics for retinal images, referring to Figure \ref{fig:figure3}. This makes the generated descriptions for retinal images more accurate and meaningful.

\section{Discussion}
\vspace{+1pt}\noindent\textbf{5.1 Challenge of Automatic Evaluation Metrics} 

In Table \ref{table:table200} and Table \ref{table:table201}, the results are based on the commonly used automatic evaluation metrics for medical report generation, e.g., BLEU \cite{papineni2002bleu}, CIDEr \cite{vedantam2015cider}, and ROUGE \cite{lin2004rouge}. Although the results show that the proposed method achieves state-of-the-art performance, scores of the automatic metrics are not that high. According to \cite{huang2021deepOpht,jing2018automatic}, this situation is quite common in medical report generation. The main reason could be that the medical image descriptions \cite{huang2021deepOpht,jing2018automatic,li2018hybrid} are much longer and complicated than the natural image descriptions \cite{xu2015show}. Hence, the innate properties \cite{papineni2002bleu,vedantam2015cider,lin2004rouge} of these automatic metrics make them suitable for natural image captioning evaluation, but not appropriate for the evaluation of medical image captioning. Using proper automatic metrics to evaluate medical image captioning models is still an open challenge  \cite{huang2021deepOpht,jing2018automatic,li2018hybrid}.

\vspace{+1pt}\noindent\textbf{5.2 Broader Impact} 

In this paper, we introduce a new medical report generation model for retinal images. Our work tries to join three different disciplines; natural language processing, computer vision, and ophthalmology \cite{jing2018automatic,huang2021deepOpht,soto2019vascular}.

According to \cite{pizzarello2004vision,huang2021deepOpht}, the World Health Organization (WHO) estimates that retinal diseases are expected to affect over 500 million people worldwide shortly. The authors of \cite{pizzarello2004vision,huang2021deepOpht} also point out that the traditional process of retinal disease diagnosis and creating a medical report for a patient takes time in practice. As we may know, deep models for computer vision or natural language processing tasks have achieved, and, in some cases, even exceeded human-level performance. The authors of \cite{huang2021deepOpht} hypothesize that an effective deep model, evaluated by commonly used automatic evaluation metrics, helps improve the conventional retinal disease treatment procedure, referring to the Figure 2 of \cite{huang2021deepOpht}, and help ophthalmologists increase diagnosis efficiency and accuracy. Base on the above hypothesis, our proposed model is an effective method to improve the traditional retinal disease treatment procedure and help ophthalmologists.

Our proposed model may have several societal benefits. Firstly, it automates part of the traditional treatment procedure, referring to the Figure 2 of \cite{huang2021deepOpht}. Hence, a member of the public/patients would need to spend less time waiting for the diagnosis information provided by retinal specialists. In addition, the diagnosis efficiency and accuracy of ophthalmologists can be improved. However, these benefits do not come without potential hazards. For example, the ability to automatically analyze retinal images and generate medical reports could also allow general users to diagnose themselves without assistance from non-retinal specialists, but the users may misunderstand the generated results. This could make them miss the golden treatment time for the potential retinal diseases. The current automatic methods can assist doctors but cannot replace them. People should have a proper/correct understanding of the usage of automatic methods.

We encourage researchers in the humanities to further investigate the ethical use and limitations of automatic medical report generation. As we may know, there is much uncertainty in medicine. An example of an important question is; What are the proper rules and regulations for using the automatic method in medicine? If some accidents happen, who should take responsibility? Doctors, patients, or method developers?

\section{Conclusion}

In conclusion, in this work, we propose a new method for context-driven medical report generation for retinal images. We also conduct various experiments to show that our proposed approach is capable of effectively leveraging the interactive information between the given context, i.e., keywords in our case, and input image. The experimental results show that our proposed method achieves state-of-the-art performance and the performance increases about $16$\% in BLEU-avg, $10.2$\% in CIDEr, and $8.6$\% in ROUGE.

\section{Acknowledgments}
This work is supported by competitive research funding from King Abdullah University of Science and Technology
(KAUST) and University of Amsterdam.

\bibliographystyle{IEEEbib}
\bibliography{strings,refs}

\end{document}